\documentclass{article}
\usepackage{ijcai20}

\usepackage{times}

\usepackage{soul}
\usepackage{url}
\usepackage[hidelinks]{hyperref}
\usepackage[utf8]{inputenc}
\usepackage[small]{caption}
\usepackage{graphicx}
\usepackage{xcolor}
\usepackage{amsmath}
\usepackage{booktabs}
\usepackage{multirow}
\usepackage{amssymb}
\usepackage{pifont}
\newcommand{\cmark}{\textcolor{black}{\ding{51}}}%
\newcommand{\xmark}{\textcolor{black}{\ding{55}}}%
\title{The Emerging Landscape of Explainable Automated Planning
\& Decision Making}

\author{
Tathagata Chakraborti$^1$\thanks{Equal contribution.} $\cdot$
Sarath Sreedharan$^{2*}$ $\cdot$
Subbarao Kambhampati$^2$
\affiliations
$^1$IBM Research AI\\
$^2$Arizona State University\\
\emails
tchakra2@ibm.com,
\{ssreedh3, rao\}@asu.edu
}

\begin{document}

\maketitle

\begin{abstract}
In this paper, we provide a comprehensive outline of the 
different threads of work in Explainable AI Planning (XAIP)
that has emerged as a focus area in the last couple of years, 
and contrast that with earlier efforts in the field in terms
of techniques, target users, and delivery mechanisms.
We hope that the survey will provide guidance to new 
researchers in automated planning towards the 
role of explanations in the effective design of 
human-in-the-loop systems, as well as provide the 
established researcher with some perspective on the 
evolution of the exciting world of explainable planning. 
\end{abstract}

\section{Introduction}

As AI techniques mature, issues of interfacing with users
has emerged as one of the primary challenges facing the 
AI community.
Primary among these challenges is for AI-based systems 
to be able to explain their reasoning to humans
in the loop \cite{gunning2019darpa}.
This is necessary both for collaborative interactions
where humans and AI systems solve problems together,
as well as in establishing trust with end users in general.
Among the work in this direction in the broader AI community, 
in this survey, we focus on how the automated planning
community in particular has responded to this challenge.


One of the recent developments towards this end
is the establishment of the Explainable AI Planning (XAIP) 
Workshop\footnote{https://kcl-planning.github.io/XAIP-Workshops/}
at the International Conference on Automated Planning 
and Scheduling (ICAPS), the premier conference in the field.
The agenda of the workshop states: 

\begin{quote}
\small \em
While XAI at large is primarily concerned with black-box learning-based approaches, model-based approaches are well suited -- arguably better suited -- for an explanation, and Explainable AI Planning (XAIP) can play an important role in helping users interface with AI technologies in complex decision-making procedures.
\end{quote}

In general, this is true for sequential decision making
tasks for a variety of reasons.
The complexity of automated planning and decision making, 
and consequently the role of explainability in it,
raises many more challenges than function approximation tasks 
(e.g. classification) 
as was originally focused on \cite{gunning2017explainable} 
by the XAI Program from DARPA.
This includes dealing with complex constraints over
problems intractable to the human's inferential capabilities,
differences in human expectations and mental models,
to proving provenance of various artifacts of a system's decision
making process over long term interactions even as the 
world evolves around it. 
Furthermore, these typically deal with reasoning tasks
where we tend to seek explanations anyway in 
human-human interactions, as opposed to perception tasks. 

Thus, the original DARPA XAI program \cite{gunning2017explainable},
which served as a great catalyst towards advancing research 
in explainable AI, has also seen evolution \cite{gunning2019darpa}
of its core focus from machine learning to the broader
sense of artificial intelligence, particularly 
decision making tasks. 
Recent surveys on the topic \cite{anjomshoae2019explainable}
also recognize this lacuna.
As the issue of explainability becomes
front and center in AI,
the importance of long term decision making
cannot be avoided \cite{wachter2017counterfactual}.
This is highlighted by the emergence of
XAI-subcommunities within planning, multi-agents,
and other communities at premier AI conferences, 
including the Explainable AI (XAI) Workshop\footnote{https://sites.google.com/view/xai2019} 
at the International Joint Conference on Artificial
Intelligence (IJCAI) 
and the Explainable Transparent 
Autonomous Agents and Multi-Agent Systems
(EXTRAAMAS) Workshop\footnote{https://extraamas.ehealth.hevs.ch} 
at the International Conference on Autonomous Agents 
and Multiagent Systems (AAMAS),
which in addition to the XAIP Workshop mentioned above, 
has captured the imagination of this
emerging field of inquiry. 

\paragraph{Survey Scope and Outline}

In this survey, we highlight the role 
of explanations in the many unique dimensions of 
a decision making problem, particularly automated
planning, and provide a comprehensive survey
of recent work in this direction.
In particular, we will focus on automated planning
as a subfield of decision making problems in order
to adhere to the limitations of a six page survey
but we will point to work in the broader area
wherever necessary to highlight themes of 
explainable planning in general.
To this end, we will start with a brief overview
of the different kinds of users associated
with a automated decision making task and
the considerations for an explanation in each case. 
We then introduce various aspects of a planning task formally 
and delve into a survey of existing works that
tackle the explanation problem in one or
more of these dimensions, while comparing and 
contrasting the properties of such explanations.
Finally, we will conclude with a summary of 
emerging trends in XAIP research. 


In the survey, we focus exclusively on explanations of a plan
as a solution of a given planning problem. 
We will not cover meta planning problems such as {\em goal
reasoning} \cite{smith2004choosing,dannenhauer2018explaining,makro}, 
or open world considerations in the explanation of plans that fail \cite{hanheide2017robot}. 
We will also not cover novel behaviors in pursuit of explainability:
e.g. the generation of {\em explicable plans} \cite{zhang2017plan} 
that conform to user expectations and are thus not required to be
explained, or the design of environments to facilitate the same \cite{xaip-sarah}. 
For a detailed treatise of the same, we refer the reader
to \cite{chakraborti2019explicability}.
Other topics excluded are execution time 
considerations, such as in \cite{langley2017explainable}.

\section{The Many Faces of XAIP}

The primary considerations in the design of explainable 
systems is the consideration of the persona of the explainee. 
This is true for explaianble AI in general \cite{zhou2020different}
but also acknowledged to be crucial to the XAIP scene as well \cite{langley2019varieties}. 


\begin{itemize}
\item[-] {\em End user:}  
This is the person who interacts with the system in the form of a user.
For a planning system, this may be the human teammate in a human-robot
team \cite{balancing} who is impacted by, or is a direct stakeholder 
in the plans of the robot, or user collaborating with an automated
planner in a decision support setting \cite{radar}.
\item[-] {\em Domain Designer:}  
This is the person involved in the acquisition of the model that the 
system works with: e.g. the designer of goal-oriented conversation systems \cite{d3wa-exp}.
\item[-] {\em Algorithm Designer:}  
The final persona is that of the developer of the algorithms
themselves: e.g. in the context of automated planning systems, 
this could be someone working on informed search.
\end{itemize}

Though \cite{langley2019varieties} does not make an explicit
distinction, for most real-world applications, 
the domain designer is distinct from the algorithm designer and may 
even not have any overlap in expertise (e.g. \cite{d3wa-exp}). 
As we go into details of different forms of XAIP techniques, 
we will see how they cater to the needs of  
one or more of these personas (c.f. Figure \ref{tab:faces}).  

\section{The Decision Making Problem}

A sequential decision making or planning problem $\Pi$
is defined in terms of a transition function 
$\delta_\Pi: A \times S \rightarrow S \times \mathbb{R}$, where
$A$ is the set of capabilities available to the agent,
$S$ is the set of states it can be in, and the real number
denotes the cost of making the transition. 
The planning algorithm $\mathbb{A}$ solves $\Pi$ subject to 
a desired property $\tau$ to produce a plan or policy $\pi$,
i.e. $\mathbb{A} : \Pi \times \tau \mapsto \pi$.
Here, $\tau$ may represent different properties such
as soundness, optimality, and so on. 

\begin{itemize}
\item {\bf Plan} $\pi = \langle a_i, a_2, \ldots, a_n\rangle, a_i \in A$ that transforms the current state $I \in S$ of the agent to its
goal $G \in S$, i.e. $\delta_\Pi(\pi, I) = \langle G, \sum_{a_i \in \pi}c_i\rangle$.
The second term in the output denotes the plan cost $c(\pi)$.
The optimal plan is $\pi^*$.
\item {\bf Policy} $\pi : s \mapsto a, a \in A, \forall\ s \in S$ 
provides a mapping from any state $s$ of the agent to the desired action $a$ to be taken in that state. The optimal policy is $\pi^*$.
\end{itemize}

While specific decision making tasks have more nuanced definitions characterizing
what forms states and actions can take, how the transition function is defined, etc.
for the purposes of this survey, this abstraction should be enough for the general audience to grasp the salient features of a decision making task and relevant XAIP concepts. 

\begin{table}[tbp!]
\centering\tiny
\begin{tabular}{@{}r|c|c|c@{}}
\toprule
 & Algorithm-based & \multicolumn{2}{c}{Model-based Explanations} \\  
 & Explanations & Inference Resolution & Model Reconciliation \\ \midrule
End User & \xmark & \cmark & \cmark \\ \midrule
Domain Designer & \xmark & \cmark & n/a \\ \midrule
Algorithm Designer & \cmark & \xmark & \xmark \\ \bottomrule
\end{tabular}
\caption{The many faces of XAIP.
}\vspace{-5pt}
\label{tab:faces}
\end{table}

\subsection{The Explanation Process}

The explanation process of a planning problem proceeds as follows, with a 
question from the explainee about the current solution of a given planning 
problem, and the explainer (the XAIP system) coming up with an 
explanation for it:

\begin{itemize}
\item[Q.]
``Why $\pi$?'' or ``Why not $\pi'$?'' 
\item[] 
Here, $\pi'$ is a {\em foil} \cite{miller2018contrastive} 
and may be either stated explicitly, implicitly,
or even partially (leading to a set of foils) in the questions.
Examples of foils would be:
\begin{itemize}
\item[-] ``Why $a \not\in \pi$?'' is a partial foil where 
all plans with action  $a$ in them are the foils. 
\item[-] The original question ``Why $\pi$?'' where the implicit foil is ``as opposed to all other plans $\pi'$''. 
\end{itemize}
\item[A.] An explanation $\mathcal{E}$ such that the explainee can compute 
$\mathbb{A} : \Pi \times \tau \mapsto \pi$ and verify that either 
\begin{itemize}
\item[] $\mathbb{A} : \Pi \times \tau \not\mapsto \pi'$; or 
\item[] $\mathbb{A} : \Pi \times \tau \mapsto \pi'$ 
but $\pi \equiv \pi'$ or $\pi > \pi'$
(the criterion for comparison may be cost, preferences, etc.).
\end{itemize}
\end{itemize}

The point of an explanation is thus to establish the property $\tau$ 
of the solution $\pi$ given a planning problem $\Pi$.
The Q\&A continues until the explainee is satisfied.
The content of an explanation $\mathcal{E}$ can
vary greatly depending on the needs of the explainee (Figure \ref{tab:faces}).
This is the topic of discussion next.

\subsection{Explanation Artifacts: Algorithm/Model/Plan}

Clearly, from the definition of the decision making task above,
there are many components at play here which can contribute to an
explanation of a plan. 
The system can explain the steps made in $\mathbb{A}$ 
while solving a problem to the debugger / algorithm designer.
It can also explain artifacts of the problem description $\Pi$
that led to the decision: these are model-based algorithm-agnostic 
explanations. These are more useful to end users.
The system can also communicate characteristics
of $\pi$ as an explanation.

It is interesting to note that this sort of a distinction 
can be seen in the literature on explainable machine learning as well. 
For example, LIME \cite{ribeiro2016should} interfaces
with the explainee at the level of outputs only, i.e.
the classification choices made (corresponding to plans
computed in our setting) -- it is also algorithm dependent 
since it reveals (albeit simplified) details of the 
learned model to the user. 
Approaches like \cite{samek2017explainable}, on the other hand,
are purely algorithm dependent requiring the explainee to visualize
the internal representations learned by the algorithm at hand. 
Other works such as \cite{datta2016algorithmic}
provide algorithm independent explanations in terms of 
the input data and black box learners, similar to
model-based explanations in our case that use the 
input problem definition as the basis of an explanation
and not the inference engine. 

\subsection{Properties of Explanations}

Existing literature on explainable artificial intelligence, as well as studies on explanations in human-human interactions, surface recurring themes used to characterize explanations. 

\paragraph{Social, Selective, and Contrastive}

Looking at how humans explain their decisions to each other
can provide great insight on the desired properties
of an explanation. 
Miller in \cite{miller2018contrastive} provides an insightful survey of lessons learned from social sciences and how they 
can impact the informed design of explainable AI systems.
He outlines three key properties for consideration: 
{\em social} in being able to model the expectations of the explainee,
{\em selective} in being able to select explanations among several 
competing hypothesis, and
{\em contrastive} in being able to differentiate
properties of two competing hypothesis.
The contrastive property in particular has received a lot of attention \cite{hoffmann2019explainable,miller2019explanation}
in the XAIP community.

\paragraph{Local versus Global Explanations}

Another consideration is whether an explanation is
geared towards a particular 
decision (local), e.g. LIME \cite{ribeiro2016should},
or they are for the entire model (global), 
e.g. TCAV \cite{kim2017interpretability} -- for a planning problem
this distinction can manifest in many ways: whether
the explanation is for a given plan versus if 
it is for the model in general.

\paragraph{Abstractions}

One final approach we want to highlight is the use of abstractions:
this is especially useful if the model of decision-making is too complex
for the explainee and a simplified model can provide more useful feedback.
\cite{ribeiro2016should}


\section{Algorithm-based Explanations}

We first look at attempts to explain the 
underlying planning algorithm.
This is quite useful for debugging: e.g.
\cite{magnaguagno2017web} provides an interactive visualization 
of the search tree for a given problem. 
Another case is where the explanation methods are particularly tailored for
specific algorithms. 
Such explanatory methods have become quite common in
explaining decisions generated by deep reinforcement learning. 
For example, authors in \cite{greydanus2018visualizing} look at the possibility of
generating perturbation based saliency maps for explaining a policy learned by
Asynchronous Advantage Actor-Critic Algorithms, while authors in \cite{koul2018learning}
look at learning finite-state representation (a Moore machine) that can represent RL
policies learned by RNNs.

\section{Model-based Explanations}

Majority of works in XAIP look at algorithm-agnostic 
methods for generating explanations since
properties of a solution can be evaluated independently of the method used to come up with it, given the model of 
the decision making task.
As opposed to debugging settings where the algorithm has to be investigated in more detail, end users typically care about model-based algorithm-agnostic explanations 
more so that services \cite{cashmore2019towards} can be built around it. 
Approaches in this category deal with two considerations: 
1) the inferential capability; and/or 2) the mental model of the user.
When both of these are aligned, there is no need to explain.

\subsection{Inference Reconciliation}

Users have considerably less computational ability
(let's say $\mathbb{A}^H$) than a planner.
In this situation:

\begin{itemize}
\item[] $\mathbb{A} : \Pi \times \tau \mapsto \pi$ and 
$\mathbb{A}^H : \Pi \times \tau \not\mapsto \pi$ 
\end{itemize}

An explanation here is supposed to reconcile the inferential power 
of the user and the planner:

\begin{itemize}
\item[] $\mathbb{A}^H : \Pi \times \tau \xrightarrow{\mathcal{E}} \pi$ 
\end{itemize}

In order to help the inference process of the user, there are usually two
broad approaches (not necessarily exclusive): (a) Allow the user to raise specific questions about a plan and engage in explanatory dialogue; and (b) leverage abstraction techniques to allow the user to better understand the plan. 

\paragraph{Investigatory Dialogue}

With a few exceptions, most of the methods that engage in explanatory dialogue look at queries contrasting the given plan with a foil (implicit or explicit). 

\begin{itemize}
\item[$Q_1$:] \em
``Why is this action in this plan or why $a \in \pi$?'' 
\end{itemize}

One of the most well-known approaches for answering this
\cite{seegebarth2012making,bercher2014plan} use a causal link chain 
originating at $a$ that can be traced to the goal.
The explanations is thus a subset of the model $\mathcal{E} \subseteq \Pi$ that
effectively explains the role of the action by pointing out the preconditions of successive actions that are being supported by the plan in question. 
While the original paper \cite{seegebarth2012making} 
does not specifically talk about any selection criterion for
the explanation content, recent work \cite{ai-comm} has shown
how the information can be minimized.

\begin{itemize}
\item[$Q_2$:] \em
``Why not this other plan $\pi'$?'' 
\end{itemize}

This is the case where a contrastive foil is explicitly considered.
Authors in \cite{cashmore2019towards,krarup2019model} assume that the foils specified by the user can be best understood as constraints on the plans
they are expecting:
e.g. a certain action/action-sequence to be included/excluded.
The explanation is then to identify an exemplary plan that satisfies those constraints thus demonstrating how the computed plan is better.
Authors in \cite{eiflernew}, on the other hand, expect the user queries to be expressed in terms of plan properties which are user-defined binary properties that apply to all valid plans for the decision problem.
The explanation then takes the form of other plan properties that are entailed by those properties. 
This is computed using oversubscription planning 
with plan properties reflecting goals.

\begin{itemize}
\item[$Q_3$:] \em 
``Why is this policy optimal, i.e. $\pi(s) = a \wedge \pi(s) \not= a'$?''
\end{itemize}


Such questions are pursued particularly in the context of MDPs:
authors in \cite{khan2009minimal} phrase explanations in terms of the frequency with which the current action would lead the agent to high-value states, while authors in
\cite{dodson_mdp_explanation} looked at such questions in a specific application context with explanations that show how the action allows for the execution of more desirable actions later. The latter additionally employs a case-based explanation technique to provide historical precedents about the results of the actions.
Authors in \cite{juozapaitis2019explainable} answer questions over $a$ being preferred over $a'$ by illustrating how the actions affect the total value in terms of various human-understandable components of the reward function.

Among these works, \cite{khan2009minimal,dodson_mdp_explanation,juozapaitis2019explainable,eiflernew} aim for minimal explanations as a means of selection. 


\begin{itemize}
\item[$Q_4$:] \em
``Why is $\Pi$ not solvable?'' 
\end{itemize}

There are several ways to surface to the user the constraints in the problem 
that are leading to unsolvability. 

\paragraph{Excuse}

One approach would be to transform the given problem to a new one so that the 
updated problem is now solvable and provide the model fix as an explanation
of why the original problem was unsolvable.

\begin{itemize}
\item[] $\Pi \rightarrow \Pi'$ so that $\mathbb{A} : \Pi' \times \tau \not\mapsto \phi$
\item[] $\mathcal{E} \leftarrow \Pi \Delta \Pi'$
\end{itemize}

These are called excuses \cite{gobelbecker2010coming}: here the authors
identify a set of static initial facts to update by framing it as a planning problem.
It is possible to impose selection strategies in this framework by associating 
costs to the various excuses.


\paragraph{Abstraction}

An alternative transformation on the problem would be to find a simpler version of the 
given problem which is still unsolvable and highlight the problems there.

\begin{itemize}
\item[] $\mathcal{E} \leftarrow Abs(\Pi)$ so that $\mathbb{A} : \mathcal{E} \times \tau \mapsto \phi$
\end{itemize}

These are called model abstractions and have been used in
\cite{sreedharan2018hierarchical,sreedharan2019can} 
to reduce the computational burden on the user.
The approach in \cite{sreedharan2019can} also leverages temporal abstractions
in the form of intermediate subgoals to illustrate why possible foils fail. 
Use of abstractions is, of course, not confined to explanations of unsolvability: 
recent work \cite{madumal2019explainable} used abstract models defined over simpler
user-defined features to generate explanations for reinforcement learning problems
in terms of action influence. 
The methods discussed in \cite{seegebarth2012making} also allows for the generation of causal link explanations for planning settings that involve abstract tasks, such as in HTN planning \cite{erol1994htn}.
The use of plan properties by \cite{eiflernew} and subsets of state factors in \cite{khan2009minimal} are more examples of the use of abstraction schemes 
to simplify the explanation process.


\paragraph{Certificates}

Finally, authors in \cite{eriksson2017unsolvability} look at a different way to
approach the unsolvability issue by creating 
inductive certificates for the initial states that captures all reachable states.
They have also investigated axiomatic systems that can generate proofs
for task unsolvability \cite{eriksson2018proof}. 
Such certificates (represented, for example, as a binary decision diagram)
can be quite complicated and are not meant to be consumed by end users,
but provide useful debugging information to domain designers, algorithm designers, and AI assistants.

\subsection{Model Reconciliation}

One of the recurring themes in human-machine interaction is
the ``mental models'' of users \cite{carroll1988mental} -- 
users of software systems often come with their own
preconceived notions and expectations of the system 
that may ore may not be borne out by the ground truth. 
For a planning system, this means that even if it is 
making the best plans it could, the human-in-the-loop
is evaluating those plans with a different model, i.e. their
mental model of the problem, and may not agree to 
its quality. Differences in models between the user and the
machine appear in many settings, such as in 
drifting world models over long terms interactions \cite{bryce2016maintaining},
search and rescue settings where there are internal and external
agents with different views into the world \cite{balancing},
in intelligent tutoring systems between the student and 
the instructor \cite{sachin},
in smart rooms with distributed sensors \cite{ai-comm}, and so on.
This model difference, along with inferential limitations of the
human, is thus the root cause of the need for explanations 
from the end user persona. 

In \cite{explain}, the seminal work on this topic, authors posit 
that explanations can no longer be a ``soliloquy`` in the 
agent's own model but must instead consider and explain 
in terms of these model differences.
The process of explanations is then one of {\em reconciliation} 
of the systems model and the human mental model so that 
both can agree on the property $\tau$ of the decision being made. 
Thus, if $\Pi^H$ is the mental model of the user,
the model reconciliation process requires that:

\begin{itemize}
\item[]
Given: $\mathbb{A} : \Pi \times \tau \mapsto \pi$
\item[] 
$\Pi^H + \mathcal{E} \rightarrow \hat{\Pi}^H$ such that 
$\mathbb{A} : \hat{\Pi}^H \times \tau \mapsto \pi$.
\end{itemize}

In the original work, the mental model was assumed to be 
known and reconciliation was achieved through a search in
the space of models induced by the difference between the 
system model and the mental model, until a model is found 
where $\tau$ holds. The difference between this intermediate 
model and the mental model is provided as an explanation.

\begin{figure*}
\centering
\includegraphics[width=\textwidth]{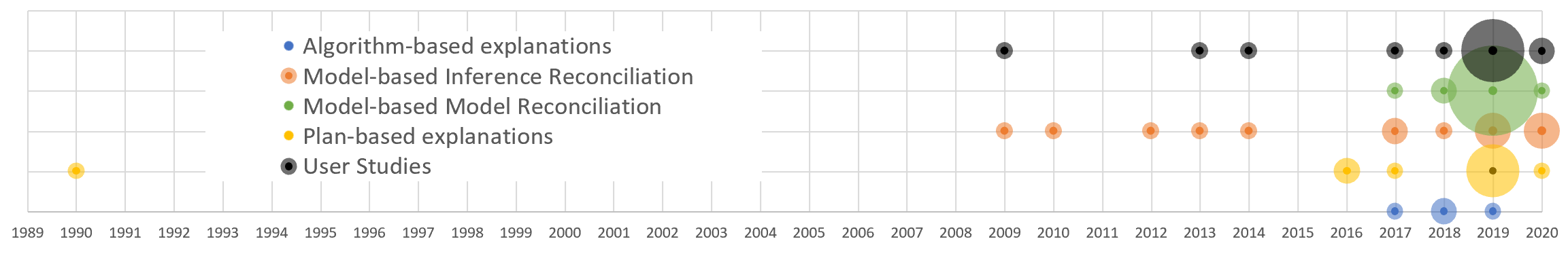}
\caption{Recent trends in XAIP illustrate 
burst in model reconciliation approaches acknowledging
the need to account for the mental model of the user
in the explanation process. 
Also noticeable is an encouraging uptick in willingness
of the community to engage in user studies.
Size of a circle is proportional to the number of papers in a year,
smallest being 1. Note that 2020 is still in progress.
}
\label{fig:trends}
\end{figure*}

\begin{table*}[tbp!]
\centering\tiny
\begin{tabular}{@{}r|r|p{1.8cm}p{2.1cm}p{1.9cm}p{2.1cm}p{1cm}p{1cm}p{1cm}@{}}
\toprule
\multicolumn{2}{c|}{Explanation Type} & Social & Contrastive & Selective & Local & Global & Abstraction & User Study \\ \midrule
\multicolumn{2}{c|}{Algorithm-based explanations} & 
\cite{juozapaitis2019explainable} & 
& 
\cite{juozapaitis2019explainable,greydanus2018visualizing,koul2018learning} & \cite{juozapaitis2019explainable,greydanus2018visualizing,magnaguagno2017web,koul2018learning} & 
& 
\cite{koul2018learning} & 
\cite{greydanus2018visualizing,magnaguagno2017web}\\ \midrule
\multirow{2}{*}{Model-based explanations} & Inference Reconciliation & 
\cite{sreedharan2018hierarchical,sreedharan2019can,eiflernew,d3wa-exp,madumal2019explainable,juozapaitis2019explainable} & 
\cite{seegebarth2012making,bercher2014plan,sreedharan2018hierarchical,sreedharan2019can,eiflernew,d3wa-exp,gobelbecker2010coming,khan2009minimal,dodson_mdp_explanation,eriksson2017unsolvability,madumal2019explainable,juozapaitis2019explainable,cashmore2019towards} & 
\cite{greydanus2018visualizing,seegebarth2012making,bercher2014plan,sreedharan2018hierarchical,sreedharan2019can,eiflernew,d3wa-exp,gobelbecker2010coming,khan2009minimal,dodson_mdp_explanation,madumal2019explainable,juozapaitis2019explainable} & 
\cite{seegebarth2012making,greydanus2018visualizing,bercher2014plan,sreedharan2018hierarchical,sreedharan2019can,eiflernew,d3wa-exp,gobelbecker2010coming,khan2009minimal,dodson_mdp_explanation,eriksson2017unsolvability,madumal2019explainable,juozapaitis2019explainable,cashmore2019towards} & 
\cite{sreedharan2019can,eiflernew,d3wa-exp,gobelbecker2010coming} & 
\cite{sreedharan2018hierarchical,sreedharan2019can,eiflernew,d3wa-exp,khan2009minimal,madumal2019explainable} & 
\cite{bercher2014plan,greydanus2018visualizing,sreedharan2019can,khan2009minimal,dodson_mdp_explanation,madumal2019explainable} \\ \cmidrule{2-9}
& Model Reconciliation & 
\cite{explain,sreedharan2018hierarchical,sreedharan2019can,d3wa-exp,sreedharan2018handling,sreedharan2019model,balancing,exact2020,ai-comm,vasileiou2019preliminary,aies-lies} & 
\cite{explain,sreedharan2018hierarchical,sreedharan2019can,d3wa-exp,sreedharan2018handling,sreedharan2019model,balancing,exact2020,ai-comm,vasileiou2019preliminary,aies-lies} &
\cite{explain,sreedharan2018hierarchical,sreedharan2019can,d3wa-exp,sreedharan2018handling,sreedharan2019model,balancing,exact2020,ai-comm,vasileiou2019preliminary,aies-lies} & 
\cite{explain,sreedharan2018hierarchical,sreedharan2018handling,sreedharan2019model,balancing,exact2020,ai-comm} & 
\cite{explain,sreedharan2019can,d3wa-exp} & 
\cite{sreedharan2018hierarchical,sreedharan2019can,d3wa-exp} & 
\cite{hri,sreedharan2019can,sreedharan2019model,balancing,aies-lies}\\ \midrule
\multicolumn{2}{c|}{Plan-based explanations} & 
\cite{offra_summ,ai-comm,rosenthal2016verbalization} & 
& 
\cite{zahavy2016graying,koul2018learning,sreedharantldr,topin2019generation,hayes2017improving,offra_summ,kim2019bayesian,ai-comm,rosenthal2016verbalization} & 
\cite{zahavy2016graying,koul2018learning,sreedharantldr,topin2019generation,hayes2017improving,offra_summ,kim2019bayesian,ai-comm,rosenthal2016verbalization,kambhampati1990classification} & 
& 
\cite{zahavy2016graying,koul2018learning,sreedharantldr,topin2019generation,hayes2017improving,rosenthal2016verbalization} & 
\cite{sreedharantldr,offra_summ} \\ \bottomrule
\end{tabular}
\caption{Summary of results. Note for reviewers: We had to go with the \texttt{plain} bib format so we can tabulate the paper numbers.}
\label{tab:summary}
\end{table*}

\paragraph{Social}

Such explanations are inherently social in being able to explicitly
capture the effect of expectations in the explanation process.
Indeed, in user studies conducted in \cite{hri}, it was shown
that participants were indeed able to identify the correct 
$\tau$ based on an explanation. 
Note that, in the model reconciliation framework, 
the mental model is just a version of the decision making problem 
at hand which the agent believes the user is operating under. 
This may be a graph, a planning problem, or even a logic program \cite{vasileiou2019preliminary}. 
The notion of model reconciliation is agnostic
to the actual representation. 

\paragraph{Contrastive}

The contrastive nature of these explanations comes 
from how the model update preserves $\tau$ of the given
plan as opposed to the foil, which may be implicitly \cite{explain}
or explicitly \cite{sreedharan2018hierarchical} provided. 
This is also closely tied with the selection process 
of those model updates. 

\paragraph{Selective}

In \cite{explain}, the explanation content was selected based
on minimality of model update: $\min | \Pi \Delta \Pi^H |$.
The minimal explanation is not unique and it was shown
in \cite{hri2} how users attribute different value to theoretically
equivalent model updates, thereby motivating further research
on how to select among several competing explanations for the user.

\subsubsection{Model Reconciliation Expansion Pack}

The last couple of years have seen extensive work on this topic.
primarily focused on relaxing the assumptions made on the 
mental model in the original model reconciliation work,
and expanding the scope of problems addressed by it. 
We expound on a few of them below.

\paragraph{Model Uncertainty}

One of the primary directions of work has been in 
considering uncertainty about the mental model. 
In \cite{sreedharan2018handling}, authors show how to 
reconcile with a set of possible mental models $\{\Pi^H_i\}$
and also demonstrate how the same framework can be used
to explain to multiple users in the loop. 
In \cite{sreedharan2018hierarchical}, on the other hand,
the authors estimate the mental model from the 
provided foil. 

\paragraph{Inference Reconciliation}

The original work on model reconciliation
assumed an user with 
identical inferential capability (optimal or sound as 
the case may be) to the planner. 
However, as we saw previously, much of XAIP has been 
about dealing with the computational limits of users. 
Model reconciliation approaches have started
adapting to this
\cite{sreedharan2018hierarchical,sreedharan2019can}
by identifying from the given foil the simplest abstraction of 
their model to explain in. 
\cite{sreedharan2019can} provides further
inferential assistance in the form of unmet subgoals.

\paragraph{Unsolvability}

An important aspect of human-planner interaction, where
inferential limitations play an outsized part, 
is the case of unsolvability. 
An interesting case of this is recently explored in \cite{d3wa-exp}
where the domain acquisition problem has been cast into the model 
reconciliation framework, reusing \cite{sreedharan2019can} to 
help out the domain designer persona when they cannot figure out
why their domain has no solutions or the solutions do not 
match their expectation.

\paragraph{Model-free Model Reconciliation} 

So far, model reconciliation has considered the mental model explicitly.
This may not be necessary.
At the end of the day, the explanation includes information regarding 
the agent model and what it include and do not include.
The mental model only helps the system to filter what new 
information is relevant to the user. 
Thus an alternative would be to predict how model information can affect
the expectation of the user \cite{sreedharan2019model}
by learning a labeling model that takes a state-action-state tuple, a subset of information about the system's model and whether the user after receiving the 
information would find this tuple explicable. 
The learned model then drives the search to determine what information 
should be exposed to the user.

\paragraph{Lies and Deception}

A consequence of going ``model-free'' is that 
the explanations provided may no longer be true
but rather be whatever users find to be satisfying. 
In the original work on model reconciliation, $\mathcal{E}$
was always constrained to be consistent with the ground truth $\Pi$.
Authors in \cite{aies-lies,xaip-lies} have shown how this constraint can be 
relaxed to hijack the model reconciliation process into producing
false explanations, opening up intriguing avenues of further 
research into the ethics of mental modeling in planning.

\section{Plan-based Explanations}

Finally, we look at the role of plans in explanatory dialogue. 
Works like \cite{sohrabi2011preferred,meadows2013seeing} have explored
explanans in the form of a plan that explains a set of observations.
Beyond human-AI interaction, the qualitative structure of plans has also been used for problems like plan-reuse and validation \cite{kambhampati1990classification}. 

\paragraph{Plan / Policy Summarization}

With regards to the role of plans in explanatory dialogue, one area we want to highlight in greater detail is that of plan or policy summarization. 
When the system is generating solutions over long time horizons 
and over large state spaces, presentation of the plan or policy to the user becomes difficult. 
One way to approach this issue is through verbalization of plans:
e.g. paths taken by a robot \cite{rosenthal2016verbalization}
along different dimensions of interest such as levels of
abstraction, specificity, and locality. 
Recent work has also attempted at domain-independent methods for plan 
summarization \cite{ai-comm} by using the model reconciliation process
with an empty mental model to compute the minimal subset of causal links
required to justify each action in a plan. 

Another possibility is to use abstraction schemes to simplify the decision structure and allow the user to drill down as required. \cite{topin2019generation} looks at the possibility of employing state abstraction that project out low importance features. 
On the other hand, \cite{sreedharantldr} generates temporal abstractions for a given policy by automatically extracting subgoals.
\cite{zahavy2016graying} takes advantage of both schemes by mapping policies learned through Deep Q-learning methods to a policy for a semi-aggregated MDP that employs both user-specified state aggregation features and temporally extended actions in the form of skills automatically generated from the learned policy. 

Another possibility would be to allow the user to ask questions about generated policies: e.g. {\em ``Under what conditions is action $a_i$ performed"?} 
This was investigated in \cite{hayes2017improving}, where both queries and answers were expressed in terms of user-specified features.
\cite{kim2019bayesian} looked at cases where the user is not just interested in learning details of the model underlying the current decisions but rather how it differs from possible alternatives,
by using LTL formulas that are true in a target set of plan traces but are not satisfied by a specified alternate set. 

A different approach is taken by \cite{offra_summ} where authors propose to present 
to users partial plans that they can figure out completions based on their knowledge 
of the task, by using various psychologically feasible computational models (e.g. models inspired by inverse reinforcement learning and imitation learning) that people could have. 






\section{Emerging Landscape}

This survey provides an overview of the 
many flavors of explainable planning and decision making
and current trends in the field.
While the works explored here are mostly
after-the-fact explanations, i.e. after a plan has been computed 
(or no plan has been found) given a planning problem,
there is recent work \cite{balancing}
demonstrating how the possibility of having to explain its decisions
can be folded into an agent's reasoning stage itself.
This is a well-known phenomenon in human behavior: we are known to make better decisions when we are asked to explain them \cite{mercier2011humans}. 
By adopting a similar philosophy, we can potentially achieve better,
more human-aware, behavior in XAIP-enabled agents as well.

Early attempts at this, employing search in the space of models \cite{balancing}, had proved computationally prohibitive.
However, recent work \cite{exact2020} has shown that 
achieving such behavior is computationally no harder 
than its classical planning counterpart! 
Furthermore, recognizing that plans are not made in vacuum but 
often in the context of interactions with end users, 
{\em can lead to a more efficient planning process
with explainable components than without}, for example,
in collaborative planning scenarios \cite{radar} or 
in anytime planners that can preserve high-level constraints
in partial plans as it plans along \cite{grea2018explainable}. 
As the XAIP community comes to terms with its
own accuracy versus efficiency trade-offs, parallel 
to similar arguments in the XAI community at large, 
a whole new world of possibilities open up 
in imbuing established planning approaches with 
the latest and best XAIP-components.

\bibliographystyle{plain}
\bibliography{bib}

\begin{thebibliography}{10}

\bibitem{anjomshoae2019explainable}
Sule Anjomshoae, Amro Najjar, Davide Calvaresi, and Kary Fr{\"a}mling.
\newblock {Explainable Agents and Robots: Results from a Systematic Literature
  Review}.
\newblock In {\em AAMAS}, 2019.

\bibitem{bercher2014plan}
Pascal Bercher, Susanne Biundo, Thomas Geier, Thilo Hoernle, Florian Nothdurft,
  Felix Richter, and Bernd Schattenberg.
\newblock {Plan, Repair, Execute, Explain -- How Planning Helps to Assemble
  Your Home Theater}.
\newblock In {\em ICAPS}, 2014.

\bibitem{bryce2016maintaining}
Dan Bryce, J~Benton, and Michael~W Boldt.
\newblock {Maintaining Evolving Domain Models}.
\newblock In {\em IJCAI}, 2016.

\bibitem{carroll1988mental}
John~M Carroll and Judith~Reitman Olson.
\newblock Mental models in human-computer interaction.
\newblock {\em Handbook of Human-Computer Interaction}, 1988.

\bibitem{cashmore2019towards}
Michael Cashmore, Anna Collins, Benjamin Krarup, Senka Krivic, Daniele
  Magazzeni, and David Smith.
\newblock {Towards Explainable AI Planning as a Service}.
\newblock In {\em ICAPS Workshop on Explainable AI Planning (XAIP)}, 2019.

\bibitem{ai-comm}
Tathagata Chakraborti, Kshitij~P. Fadnis, Kartik Talamadupula, Mishal Dholakia,
  Biplav Srivastava, Jeffrey~O. Kephart, and Rachel K.~E. Bellamy.
\newblock {Planning and Visualization for a Smart Meeting Room Assistant -- A
  Case Study in the Cognitive Environments Laboratory at IBM T.J. Watson
  Research Center, Yorktown}.
\newblock {\em AI Communication}, 2019.

\bibitem{xaip-lies}
Tathagata Chakraborti and Subbarao Kambhampati.
\newblock {(How) Can AI Bots Lie?}
\newblock In {\em ICAPS Workshop on Explainable AI Planning (XAIP)}, 2019.

\bibitem{aies-lies}
Tathagata Chakraborti and Subbarao Kambhampati.
\newblock {(When) Can AI Bots Lie?}
\newblock In {\em AIES/AAAI}, 2019.

\bibitem{chakraborti2019explicability}
Tathagata Chakraborti, Anagha Kulkarni, Sarath Sreedharan, David~E Smith, and
  Subbarao Kambhampati.
\newblock {Explicability? Legibility? Predictability? Transparency? Privacy?
  Security? The Emerging Landscape of Interpretable Agent Behavior}.
\newblock In {\em ICAPS}, 2019.

\bibitem{hri}
Tathagata Chakraborti, Sarath Sreedharan, Sachin Grover, and Subbarao
  Kambhampati.
\newblock {Plan Explanations as Model Reconciliation -- An Empirical Study}.
\newblock In {\em HRI}, 2019.

\bibitem{balancing}
Tathagata Chakraborti, Sarath Sreedharan, and Subbarao Kambhampati.
\newblock {Balancing Explanations and Explicability in Human-Aware Planning}.
\newblock In {\em IJCAI}, 2019.

\bibitem{explain}
Tathagata Chakraborti, Sarath Sreedharan, Yu~Zhang, and Subbarao Kambhampati.
\newblock {Plan Explanations as Model Reconciliation: Moving Beyond Explanation
  as Soliloquy}.
\newblock In {\em IJCAI}, 2017.

\bibitem{dannenhauer2018explaining}
Dustin Dannenhauer, Michael~W Floyd, Daniele Magazzeni, and David~W Aha.
\newblock {Explaining Rebel Behavior in Goal Reasoning Agents}.
\newblock In {\em ICAPS Workshop on Explainable AI Planning (XAIP)}, 2018.

\bibitem{datta2016algorithmic}
Anupam Datta, Shayak Sen, and Yair Zick.
\newblock Algorithmic transparency via quantitative input influence: Theory and
  experiments with learning systems.
\newblock In {\em IEEE Symposium on Security and Privacy (SP)}, 2016.

\bibitem{dodson_mdp_explanation}
Thomas Dodson, Nicholas Mattei, Joshua~T. Guerin, and Judy Goldsmith.
\newblock {An English-Language Argumentation Interface for Explanation
  Generation with Markov Decision Processes in the Domain of Academic
  Advising}.
\newblock {\em TiiS}, 2013.

\bibitem{eiflernew}
Rebecca Eifler, Michael Cashmore, J{\"o}rg Hoffmann, Daniele Magazzeni, and
  Marcel Steinmetz.
\newblock {A New Approach to Plan-Space Explanation: Analyzing Plan-Property
  Dependencies in Oversubscription Planning}.
\newblock In {\em AAAI}, 2020.

\bibitem{eriksson2017unsolvability}
Salom{\'e} Eriksson, Gabriele R{\"o}ger, and Malte Helmert.
\newblock {Unsolvability Certificates for Classical Planning}.
\newblock In {\em ICAPS}, 2017.

\bibitem{eriksson2018proof}
Salom{\'e} Eriksson, Gabriele R{\"o}ger, and Malte Helmert.
\newblock {A Proof System for Unsolvable Planning Tasks}.
\newblock In {\em ICAPS}, 2018.

\bibitem{erol1994htn}
Kutluhan Erol, James Hendler, and Dana~S Nau.
\newblock {HTN planning: Complexity and Expressivity}.
\newblock In {\em AAAI}, 1994.

\bibitem{gobelbecker2010coming}
Moritz G{\"o}belbecker, Thomas Keller, Patrick Eyerich, Michael Brenner, and
  Bernhard Nebel.
\newblock {Coming Up with Good Excuses: What to Do When No Plan Can be Found}.
\newblock In {\em ICAPS}, 2010.

\bibitem{grea2018explainable}
Antoine Grea, La{\"e}titia Matignon, and Samir Aknine.
\newblock {How Explainable Plans Can Make Planning Faster}.
\newblock In {\em IJCAI Workshop on Explainable AI (XAI)}, 2018.

\bibitem{greydanus2018visualizing}
Samuel Greydanus, Anurag Koul, Jonathan Dodge, and Alan Fern.
\newblock {Visualizing and Understanding Atari Agents}.
\newblock In {\em ICML}, 2018.

\bibitem{sachin}
Sachin Grover, Tathagata Chakraborti, and Subbarao Kambhampati.
\newblock {What Can Automated Planning do for Intelligent Tutoring Systems?}
\newblock In {\em ICAPS Scheduling and Planning Applications Workshop (SPARK)},
  2018.

\bibitem{radar}
Sachin Grover, Sailik Sengupta, Tathagata Chakraborti, Aditya~Prasad Mishra,
  and Subbarao Kambhampati.
\newblock {RADAR: Automated Task Planning for Proactive Decision Support}.
\newblock {\em HCI Journal}, 2020.

\bibitem{gunning2017explainable}
David Gunning.
\newblock {Explainable Artificial Intelligence (XAI)}.
\newblock {\em Defense Advanced Research Projects Agency (DARPA)}, 2017.

\bibitem{gunning2019darpa}
David Gunning and David~W Aha.
\newblock {DARPA's Explainable Artificial Intelligence Program}.
\newblock {\em AI Magazine}, 2019.

\bibitem{hanheide2017robot}
Marc Hanheide, Moritz G{\"o}belbecker, Graham~S Horn, Andrzej Pronobis,
  Kristoffer Sj{\"o}{\"o}, Alper Aydemir, Patric Jensfelt, Charles Gretton,
  Richard Dearden, Miroslav Janicek, et~al.
\newblock {Robot Task Planning and Explanation in Open and Uncertain Worlds}.
\newblock {\em Artificial Intelligence}, 2017.

\bibitem{hayes2017improving}
Bradley Hayes and Julie~A Shah.
\newblock {Improving Robot Controller Transparency Through Autonomous Policy
  Explanation}.
\newblock In {\em HRI}, 2017.

\bibitem{hoffmann2019explainable}
J{\"o}rg Hoffmann and Daniele Magazzeni.
\newblock {Explainable AI Planning (XAIP): Overview and the Case of Contrastive
  Explanation}.
\newblock In {\em Reasoning Web. Explainable Artificial Intelligence}, 2019.
\newblock Extended Abstract.

\bibitem{juozapaitis2019explainable}
Zoe Juozapaitis, Anurag Koul, Alan Fern, Martin Erwig, and Finale Doshi-Velez.
\newblock {Explainable Reinforcement Learning via Reward Decomposition}.
\newblock In {\em IJCAI Workshop on Explainable AI (XAI)}, 2019.

\bibitem{kambhampati1990classification}
Subbarao Kambhampati.
\newblock {A Classification of Plan Modification Strategies Based on Coverage
  and Information Requirements}.
\newblock In {\em AAAI Spring Symposium on Case Based Reasoning}, 1990.

\bibitem{khan2009minimal}
Omar~Zia Khan, Pascal Poupart, and James~P Black.
\newblock {Minimal Sufficient Explanations for Factored Markov Decision
  Processes}.
\newblock In {\em ICAPS}, 2009.

\bibitem{kim2017interpretability}
Been Kim, Martin Wattenberg, Justin Gilmer, Carrie Cai, James Wexler, Fernanda
  Viegas, and Rory Sayres.
\newblock {Interpretability Beyond Feature Attribution: Quantitative Testing
  with Concept Activation Vectors (TCAV)}.
\newblock In {\em ICML}, 2018.

\bibitem{kim2019bayesian}
Joseph Kim, Christian Muise, Ankit Shah, Shubham Agarwal, and Julie Shah.
\newblock {Bayesian Inference of Linear Temporal Logic Specifications for
  Contrastive Explanations}.
\newblock In {\em IJCAI}, 2019.

\bibitem{koul2018learning}
Anurag Koul, Sam Greydanus, and Alan Fern.
\newblock {Learning Finite State Representations of Recurrent Policy Networks}.
\newblock In {\em ICLR}, 2018.

\bibitem{krarup2019model}
Benjamin Krarup, Michael Cashmore, Daniele Magazzeni, and Tim Miller.
\newblock {Model-Based Contrastive Explanations for Explainable Planning}.
\newblock In {\em ICAPS Workshop on Explainable AI Planning (XAIP)}, 2019.

\bibitem{xaip-sarah}
Anagha Kulkarni, Sarath Sreedharan, Sarah Keren, Tathagata Chakraborti,
  David~E. Smith, and Subbarao Kambhampati.
\newblock {Design for Interpretability}.
\newblock {\em ICAPS Workshop on Explainable AI Planning (XAIP)}, 2019.

\bibitem{offra_summ}
Isaac Lage, Daphna Lifschitz, Finale Doshi-Velez, and Ofra Amir.
\newblock Exploring computational user models for agent policy summarization.
\newblock In {\em IJCAI}, 2019.

\bibitem{langley2019varieties}
Pat Langley.
\newblock Varieties of explainable agency.
\newblock In {\em ICAPS Workshop on Explainable AI Planning (XAIP)}, 2019.

\bibitem{langley2017explainable}
Pat Langley, Ben Meadows, Mohan Sridharan, and Dongkyu Choi.
\newblock {Explainable Agency for Intelligent Autonomous Systems}.
\newblock In {\em IAAI/AAAI}, 2017.

\bibitem{madumal2019explainable}
Prashan Madumal, Tim Miller, Liz Sonenberg, and Frank Vetere.
\newblock {Explainable Reinforcement Learning Through a Causal Lens}.
\newblock In {\em AAAI}, 2020.

\bibitem{magnaguagno2017web}
Maur{\i}cio~C Magnaguagno, Ramon~Fraga Pereira, Martin~D M{\'o}re, and Felipe
  Meneguzzi.
\newblock {Web Planner: A Tool to Develop Classical Planning Domains and
  Visualize Heuristic State-Space Search}.
\newblock In {\em ICAPS Workshop on User Interfaces in Scheduling and Planning
  (UISP)}, 2017.

\bibitem{makro}
Raymond Sheh David Aha Piyabutra Jampathom Keith Akins Eric Sydow
  Vikas~Shivashankar Mark~Roberts, Isaac~Monteath and Claude Sammut.
\newblock {What was I planning to do?}
\newblock In {\em ICAPS Workshop on Explainable AI Planning (XAIP)}, 2018.

\bibitem{meadows2013seeing}
Ben~Leon Meadows, Pat Langley, and Miranda~Jane Emery.
\newblock {Seeing Beyond Shadows: Incremental Abductive Reasoning for Plan
  Understanding}.
\newblock In {\em AAAI Workshop on Plan, Activity, and Intent Recognition
  (PAIR)}, 2013.

\bibitem{mercier2011humans}
Hugo Mercier and Dan Sperber.
\newblock {Why do Humans Reason? Arguments for an Argumentative Theory}.
\newblock {\em Behavioral and Brain Sciences}, 2011.

\bibitem{miller2018contrastive}
Tim Miller.
\newblock {Contrastive Explanation: A Structural-Model Approach}.
\newblock {\em arXiv:1811.03163}, 2018.

\bibitem{miller2019explanation}
Tim Miller.
\newblock {Explanation in Artificial Intelligence: Insights from the Social
  Sciences}.
\newblock {\em Artificial Intelligence}, 2019.

\bibitem{ribeiro2016should}
Marco~Tulio Ribeiro, Sameer Singh, and Carlos Guestrin.
\newblock {``Why Should I Trust You?'' Explaining the Predictions of Any
  Classifier}.
\newblock In {\em KDD}, 2016.

\bibitem{rosenthal2016verbalization}
Stephanie Rosenthal, Sai~P Selvaraj, and Manuela~M Veloso.
\newblock {Verbalization: Narration of Autonomous Robot Experience}.
\newblock In {\em IJCAI}, 2016.

\bibitem{samek2017explainable}
Wojciech Samek, Thomas Wiegand, and Klaus-Robert M{\"u}ller.
\newblock {Explainable Artificial Intelligence: Understanding, Visualizing and
  Interpreting Deep Learning Models}.
\newblock {\em arXiv:1708.08296}, 2017.

\bibitem{seegebarth2012making}
Bastian Seegebarth, Felix M{\"u}ller, Bernd Schattenberg, and Susanne Biundo.
\newblock {Making Hybrid Plans More Clear to Human Users -- A Formal Approach
  for Generating Sound Explanations}.
\newblock In {\em ICAPS}, 2012.

\bibitem{smith2004choosing}
David~E Smith.
\newblock {Choosing Objectives in Over-Subscription Planning}.
\newblock In {\em ICAPS}, 2004.

\bibitem{sohrabi2011preferred}
Shirin Sohrabi, Jorge~A Baier, and Sheila~A McIlraith.
\newblock {Preferred Explanations: Theory and Generation via Planning}.
\newblock In {\em AAAI}, 2011.

\bibitem{exact2020}
Sarath Sreedharan, Tathagata Chakraborti, Christian Muise, and Subbarao
  Kambhampati.
\newblock {Expectation-Aware Planning: A Unifying Framework for Synthesizing
  and Executing Self-Explaining Plans for Human-Aware Planning}.
\newblock In {\em AAAI}, 2020.

\bibitem{d3wa-exp}
Sarath Sreedharan, Tathagata Chakraborti, Christian Muise, Yasaman Khazaeni,
  and Subbarao Kambhampati.
\newblock {D3WA+: A Case Study of XAIP in a Model Acquisition Task}.
\newblock In {\em ICAPS}, 2020.

\bibitem{sreedharan2019model}
Sarath Sreedharan, Alberto~Olmo Hernandez, Aditya~Prasad Mishra, and Subbarao
  Kambhampati.
\newblock {Model-Free Model Reconciliation}.
\newblock In {\em IJCAI}, 2019.

\bibitem{sreedharan2018handling}
Sarath Sreedharan, Subbarao Kambhampati, et~al.
\newblock {Handling Model Uncertainty and Multiplicity in Explanations via
  Model Reconciliation}.
\newblock In {\em ICAPS}, 2018.

\bibitem{sreedharan2018hierarchical}
Sarath Sreedharan, Siddharth Srivastava, and Subbarao Kambhampati.
\newblock {Hierarchical Expertise Level Modeling for User Specific Contrastive
  Explanations}.
\newblock In {\em IJCAI}, 2018.

\bibitem{sreedharantldr}
Sarath Sreedharan, Siddharth Srivastava, and Subbarao Kambhampati.
\newblock {TLdR: Policy Summarization for Factored SSP Problems Using Temporal
  Abstractions}.
\newblock In {\em ICAPS}, 2020.

\bibitem{sreedharan2019can}
Sarath Sreedharan, Siddharth Srivastava, David Smith, and Subbarao Kambhampati.
\newblock {Why Can’t You Do That HAL? Explaining Unsolvability of Planning
  Tasks}.
\newblock In {\em IJCAI}, 2019.

\bibitem{topin2019generation}
Nicholay Topin and Manuela Veloso.
\newblock {Generation of Policy-Level Explanations for Reinforcement Learning}.
\newblock In {\em AAAI}, 2019.

\bibitem{vasileiou2019preliminary}
Stylianos Vasileiou, William Yeoh, and Tran~Cao Son.
\newblock {A Preliminary Logic-based Approach for Explanation Generation}.
\newblock In {\em ICAPS Workshop on Explainable AI Planning (XAIP)}, 2019.

\bibitem{wachter2017counterfactual}
Sandra Wachter, Brent Mittelstadt, and Chris Russell.
\newblock {Counterfactual Explanations Without Opening the Black Box: Automated
  Decisions and the GDPR}.
\newblock {\em Harvard Journal of Law \& Technology}, 2017.

\bibitem{zahavy2016graying}
Tom Zahavy, Nir Ben-Zrihem, and Shie Mannor.
\newblock {Graying the Black Box: Understanding DQNs}.
\newblock In {\em ICML}, 2016.

\bibitem{hri2}
Zahra Zahedi, Alberto Olmo, Tathagata Chakraborti, Sarath Sreedharan, and
  Subbarao Kambhampati.
\newblock {Towards Understanding User Preferences for Explanation Types in
  Explanation as Model Reconciliation}.
\newblock In {\em HRI}, 2019.
\newblock Late Breaking Report.

\bibitem{zhang2017plan}
Yu~Zhang, Sarath Sreedharan, Anagha Kulkarni, Tathagata Chakraborti,
  Hankz~Hankui Zhuo, and Subbarao Kambhampati.
\newblock {Plan Explicability and Predictability for Robot Task Planning}.
\newblock In {\em ICRA}, 2017.

\bibitem{zhou2020different}
Yishan Zhou and David Danks.
\newblock {Different ``Intelligibility'' for Different Folks}.
\newblock In {\em AIES/AAAI}, 2020.

\end{thebibliography}

\end{document}